\def\year{2018}\relax
\documentclass[letterpaper]{article} %DO NOT CHANGE THIS
\usepackage{aaai18}  %Required
\usepackage{times}  %Required
\usepackage{helvet}  %Required
\usepackage{courier}  %Required
\usepackage{url}  %Required
\usepackage{graphicx,subcaption}  %Required
\usepackage{amsmath,amsthm,amssymb}
\usepackage{mathtools}
\usepackage{mathrsfs}
\usepackage{enumitem}
\usepackage{multirow}
\usepackage[sort, numbers, square]{natbib}
\usepackage[table,xcdraw]{xcolor}
\usepackage{colortbl}
\usepackage{multirow}
\usepackage{algorithm}
\usepackage{algorithmic}
\usepackage{float}
\usepackage{dsfont}

\theoremstyle{definition}
\newtheorem{definition}{Definition}

\begin{document}
\title{Verification of Recurrent Neural Networks Through Rule Extraction\thanks{Some of the work discussed here can be found at~\cite{Empirical2017Wang} and~\cite{Comparison2018Wang}}}
\author
{Qinglong Wang,$^{1}$ Kaixuan Zhang,$^{2}$ Xue Liu,$^{1}$ C. Lee Giles,$^{2}$\\
\\
\normalsize{$^{1}$School of Computer Science, McGill University, Canada}\\
%\normalsize{An Unknown Address, Wherever, ST 00000, USA}\\
\normalsize{$^{2}$Information Sciences and Technology, Pennsylvania State University, USA}\\
\\
%\normalsize{$^\ast$To whom correspondence should be addressed; Email: giles@ist.psu.edu}
}

\maketitle
\begin{abstract}
The verification problem for neural networks is verifying whether a neural network will suffer from adversarial samples, or approximating the maximal allowed scale of adversarial perturbation that can be endured. While most prior work contributes to verifying feed-forward networks, little has been explored for verifying recurrent networks. This is due to the existence of a more rigorous constraint on the perturbation space for sequential data, and the lack of a proper metric for measuring the perturbation. In this work, we address these challenges by proposing a metric which measures the distance between strings, and use deterministic finite automata (DFA) to represent a rigorous oracle which examines if the generated adversarial samples violate certain constraints on a perturbation. More specifically, we empirically show that certain recurrent networks allow relatively stable DFA extraction. As such, DFAs extracted from these recurrent networks can serve as a surrogate oracle for when the ground truth DFA is unknown. We apply our verification mechanism to several widely used recurrent networks on a set of the Tomita grammars. The results demonstrate that only a few models remain robust against adversarial samples. In addition, we show that for grammars with different levels of complexity, there is also a difference in the difficulty of robust learning of these grammars. 
\end{abstract}
\section{Introduction}
\label{sec:intro}
Verification for neural networks is crucial for validating deep learning techniques in security critical applications. However, the black-box nature of neural networks makes inspection, analysis, and verification of their captured knowledge difficult or near-impossible~\cite{omlin2000symbolic}. Moreover, the complicated architecture of neural networks also make these models vulnerable to adversarial attacks~\cite{SzegedyZSBEGF13} -- a synthetic sample generated by slightly modifying a source sample in order to trick a neural network into ``believing'' this modified sample belongs to an incorrect class with high confidence.

Most prior work on neural network verification has been on verifying feed-forward neural networks using mixed-integer linear programming (MILP)~\cite{tjeng2017evaluating,ChengNR17, Fischetti17milp,MadryMSTV17} and Satisfiability Modulo Theories (SMT)~\cite{Carlini17Ground,Ehlers17,KatzBDJK17}. Specifically, these approaches can either verify if a neural network can remain robust to a constrained perturbation applied to an input, or approximate the maximal allowed scale of the perturbation that can be tolerated. To apply these verification approaches, two critical requirements need to be satisfied. One is that an adversarial sample should be recognized by a hypothetical oracle that is very similar or even identical to its source sample. Another requirement is that the adversarial perturbation must be of small enough scale to avoid being detected by the oracle.

Depending on applications, there are different ways to set up the oracle and various distance metrics that measure the scale of an adversarial perturbation. For image recognition, fortunately, it is not challenging to satisfy the two requirements mentioned above. More specifically, in this scenario, a human is usually assumed to be the oracle and adversarial images must avoid a straightforward visual inspection. However, it is neither realistic nor efficient to assign a human oracle. As such, the oracle in this case can be simply replaced by the ground truth labels of source images. As for the distance metrics, various $L_p$ norms ($p = 0, 2, \infty$) have been widely adopted in prior work~\cite{SzegedyZSBEGF13,WangGZOXLG17,Carlini17Ground} on adversarial sample problem. The convenience brought by image recognition has made this application as the benchmark for much verification work~\cite{tjeng2017evaluating,fischetti2017deep,ehlers2017formal,narodytska2017verifying,kevorchian2018verification}. 

When dealing with sequential data, e.g. natural language, programming code, DNA sequence, etc., however these requirements are challenging to satisfy. This is mainly due to the lack of appropriate oracles and proper distance metrics. For instance in sentiment analysis, it has been shown that even the change of a single word is sufficient to fool a recurrent neural network (RNN)~\cite{PapernotMSH16}. However, the adversarial sentence presented in this work~\cite{PapernotMSH16} contains grammatical errors. This indicates that for sequential data, the adversarial samples need not only be negligible, but also satisfy certain grammatical or semantic constraints. Unfortunately, it is very challenging to formulate these constraints and construct an oracle with these constraints. Since RNNs are often used for processing sequential data, the difficulty of verifying sequential data has consequents which limits research work on verifying RNNs.

Here, we propose to use deterministic finite automata (DFA) as the oracle. There exists much prior work on relating RNNs to DFA. Our line of research aims at extracting rules from RNNs, where extracted rules are usually expressed by a DFA. Furthermore, we design a distance metric -- average edit distance -- for measuring and constraining the perturbations applied to strings generated by regular grammars. Since it is very difficult, if not impossible, to design comprehensive distance metrics for real-world sequential data, we propose this work as a steppingstone for verifying RNNs that can be built for more sophisticated applications and can have extracted rules. In summary, this work makes the following contributions:

\begin{itemize}
\item We propose a distance metric for measuring the scale of adversarial perturbations applied to strings generated by regular grammars. We show that the average edit distance can also describe the complexity of regular grammars.
\item We empirically study the factors that influence DFA extraction, and conduct a careful experimental study of evaluating and comparing different recurrent networks for DFA extraction on the Tomita grammars~\cite{tomita1982}. Our results show that, despite these factors, DFA can be stably extracted from second-order RNNs~\cite{giles1990higher}. In addition, among all RNNs investigated, RNNs with strong quadratic (or approximate quadratic) forms of hidden layer interaction provide the most accurate and stable DFA extraction for all of the Tomita grammars.
\item We demonstrate that using DFA can evaluate the adversarial accuracy of different RNNs on Tomita grammars. The experiments show the difference between the robustness of RNNs and the difference in the difficulty of robust learning of grammars with different complexity.
\end{itemize}

\section{Verification Framework for RNNs}
\label{sec:overview}
The verification problem for neural networks is typically formulated as a MILP or SMT problem. Our work is closely related to prior work on verifying feed-forward neural networks~\cite{Fischetti17milp, tjeng2017evaluating} and propose the following formulation for verifying recurrent networks.

We first denote the domain of all regular strings as $\mathcal{X} = \Sigma^{*}$, where $\Sigma$ is the alphabet for regular strings. Then we denote the oracle by $\lambda$, which can process any $x \in \mathcal{X}$ and produce a classification decision $\lambda(x) \in \mathcal{Y}$. The set of strings classified by $\lambda$ having the same label of $y$ is denoted by $\mathcal{X}_y$, i.e. $\mathcal{X}_y = \{ x \mid \lambda(x) = y \}$. We assume there is a distance metric denoted by $d$ (detailed introduction of our defined distance metric is provided in Section~\ref{sec:dist}.) to measure the distance between strings. Let $\mathcal{P}(x)$ denote the set of all possible strings generated by perturbing $x$ with respect to a certain distance constraint, i.e. $\mathcal{P}(x) = \{ x' \mid d (x, x') \leq d (x, \mathcal{X}_{\mu})\}$, where $\mu \in \mathcal{Y}\setminus  \lambda(x)$, represents any label that is different from $\lambda(x)$. The above constraint indicates that the allowed perturbation to $x$ must not lead to a different classification result made by $\lambda$. 

Similarly, a RNN $f$ can process any $x \in \mathcal{X}$ and produce a vector of classification scores, i.e. $f(x) \in \mathbb{R}^{\mid \mathcal{Y} \mid}$ and $\sum_{i=1}^{\mid \mathcal{Y} \mid} f_{i}(x) = 1$. Then we say $f$ is robust or locally invariant~\cite{leofante2018automated} with respect to $\hat{x}$ if and only if finding a $x'$ that satisfying~\eqref{eq:invariance} is infeasible:
\begin{equation}
  \begin{aligned}
  \label{eq:invariance}
  \big (x' \!\in\! \mathcal{X} \cap \mathcal{P}(\hat{x}) \cap  \mathcal{X}_{\lambda(\hat{x})} \big ) \!\wedge\! \big (f_{\lambda(\hat{x})}(x') < \!\! \underset{\mu \in \mathcal{Y} \setminus \lambda(\hat{x})}{\mathrm{max}}\! f_{\mu}(x') \big)
  \end{aligned}
\end{equation}

To describe the relation between $f$ and $\lambda$ from a global perspective, we adapt the local invariance property described above to determine the equivalence~\cite{leofante2018automated} between $f$ and $\lambda$. More formally, we say there exists an equivalence relation between $f$ and $\lambda$ if it is infeasible to find a $x$ that satisfies the following:
\begin{equation}
  \begin{aligned}
  \label{eq:equivalence}
  \big (x \!\in\! \mathcal{X} ) \!\wedge\! \big (\mathrm{arg}\,\underset{\mu \in \mathcal{Y}}{\mathrm{max}} \; f_{\mu}(x) \neq \lambda(x) \big).
  \end{aligned}
\end{equation}

As discussed in Section~\ref{sec:intro}, $\lambda$ and $d$ play two crucial roles in our verification framework.  As such, it is important to have some $\lambda$ with high ``quality'' to represent the oracle. Our prior work~\cite{Empirical2017Wang} demonstrated that for certain RNNs, DFAs with high classification accuracy can be extracted in a relatively stable manner. As such, we use DFAs as oracles for verifying recurrent networks. In addition, it needs to be noted that equation~\eqref{eq:equivalence} provides a way to evaluate the fidelity (introduced in Section~\ref{sec:extract_metrics}) of an extracted DFA regarding its source RNN. This is important since analyzing an extracted DFA with high fidelity can be more tractable than analyzing its complicated source RNN. Also, our previous work~\cite{Comparison2018Wang} defined the \emph{average edit distance} that measures the difference between different sets of strings. From this, it appears that the average edit distance can also be applied to our verification framework. In the following sections, we will introduce the average edit distance, followed by our empirical study on extracting DFAs from various RNNs, and verifying these RNNs with DFAs.

\section{Average Edit Distance}
\label{sec:dist}
Our definition of average edit distance is an extension of the common definition of \emph{edit} distance, which measures the minimum number of operations -- insertion, deletion and substitution of one symbol from a string -- needed to covert a string into another~\cite{de2010grammatical}. We use this definition for the problem of measuring the difference between sets of strings. One particular application of this metric is for evaluating the complexity of regular grammars~\cite{Comparison2018Wang}, where we measure the difference between the sets of strings that are accepted and rejected by a regular grammar.
.

\subsection{Definition of Average Edit Distance}
\label{sec:def_dist}
Without loss of generality, we consider for simplicity only two sets of strings with different labels. Given a string $x \in X_{y}^N$ and a string $x' \in X_{y'}^N$, where $X_{y}^N$ and $X_{y'}^N$ denote the sets of strings with length $N$ and labels $y$ and $y'$, respectively. The edit distance between $x$ and the set of strings in $X_{y'}^N$ can be expressed as: 
\begin{equation}
  \begin{aligned}
  \label{eq:x_avg_edit_dist}
  \bar{d}_{e}(x, X_{y'}^N) &= \underset{x' \in X_{y'}^N}{\mathrm{min}} d_{e}(x, x').
  \end{aligned}
\end{equation}
We then have the following definition.
\begin{definition}[Average Edit Distance]
\label{def:ave_edit_dist}
The average edit distance $D(\mathcal{X}_{y}, \mathcal{X}_{y'})$ between two sets of strings $\mathcal{X}_{y}$ and $\mathcal{X}_{y'}$ is: 
\begin{equation}
  \begin{aligned}
  \label{eq:ave_edit_dist}
  D(\mathcal{X}_{y}, \mathcal{X}_{y'}) = \frac{1}{2} \cdot \underset{N \rightarrow \infty }{\mathrm{lim}}  \big (\frac{1}{|X_{y}^N|} D_{y}^{N} + \frac{1}{|X_{y'}^N|} D_{y'}^{N} \big ),
  \end{aligned}
  %\vspace{-0.5em}
\end{equation}
where $D_{y}^{N}$ and $D_{y'}^{N}$ denote $\sum_{x \in X_{y}^N} \bar{d}_{e}(x, X_{y'}^N)$ and $\sum_{x' \in X_{y'}^N} \bar{d}_{e}(x', X_{y}^N)$, respectively.
\end{definition}

If we let $\mathcal{X}_{y}$ and $\mathcal{X}_{y'}$ represent the sets of accepted and rejected strings for a certain regular grammar, then $D(\mathcal{X}_{y}, \mathcal{X}_{y'})$ essentially reflects the complexity of this grammar~\cite{Comparison2018Wang}. In particular, a grammar with higher complexity (hence smaller average edit distance) will be more challenging for robust learning (we show this result in Section~\ref{sec:exp}). In the following, we show the case of how to use the average edit distance to categorize a certain set of regular grammars.

\begin{table}[t]
\small
\centering
  \begin{tabular}{cl}
  \hline \hline
  G & Description\\                                \hline \hline
  1 & $1^{*}$    \\                                       \hline
  2 & $(1 0)^{*}$ \\                                      \hline
  3 & \begin{tabular}[c]{@{}l@{}}an odd number of consecutive 1s is always followed \\by an even number of consecutive 0s
      \end{tabular}                                    \\ \hline
  4 & any string not containing ``000'' as a substring \\ \hline
  5 & even number of 0s and even number of 1s~\cite{giles1990higher}                              \\ \hline
  6 & \begin{tabular}[c]{@{}l@{}}the difference between the number of 0s and the \\number of 1s is a multiple of 3
      \end{tabular}                                    \\ \hline
  7 & $0^{*}1^{*}0^{*}1^{*}$                           \\ \hline \hline
  \end{tabular}
  \caption{Descriptions of Tomita grammars.}
  \label{tab:tomita}
\end{table}

\subsection{Average Edit Distance for Tomita Grammars}
\label{sec:tomita_dist}
Tomita grammars~\cite{tomita1982} denote a set of seven regular grammars and have been widely adopted in the study of DFA extraction for recurrent networks. These grammars all have alphabet $\Sigma = \{0,1\}$, and generate an infinite language over $\{0,1\}^{*}$. A description of the Tomita grammars is provided in Table~\ref{tab:tomita}. For a more detailed introduction of Tomita grammars, please see Tomita's early work~\cite{tomita1982}. 
\begin{table}[t]
\centering
\small
\caption{Average edit distance for Tomita grammars.}
\label{tab:computed_dist}
\begin{tabular}{ccccccccc}
\hline\hline
\multirow{2}{*}{ } & \multirow{2}{*}{$N$} & \multicolumn{7}{c}{Grammar}                    \\ \cline{3-9} 
                      &                       & G1   & G2   & G3   & G4   & G5   & G6   & G7   \\ \hline\hline
\multirow{4}{*}{$D^{N}$} & 8                   & 2.51 & 2.51 & 1.13 & 1.16 & 1.00 & 1.00 & 1.17 \\
                      & 10                  & 3.00 & 3.00 & 1.18 & 1.16 & 1.00 & 1.00 & 1.31 \\
                      & 12                  & 3.50 & 3.50 & 1.24 & 1.18 & 1.00 & 1.00 & 1.51 \\
                      & 14                  & 4.00 & 4.00 & 1.30 & 1.22 & 1.00 & 1.00 & 1.75 \\ \hline\hline
\end{tabular}
\end{table}

Using Definition~\ref{def:ave_edit_dist}, we can calculate the average edit distance for the Tomita grammars. As shown in Table~\ref{tab:computed_dist}, different Tomita grammars have different values and changing trends of average edit distance as we increase the length of strings. More specifically, as $N$ increases, the average edit distance of grammars 1, 2 and 7 monotonically increases, while for other grammars, their average edit distance increases at a slower rate (grammar 3, 4) or remain constant (grammar 5, 6). These observations allow us to categorize Tomita grammars into the following three classes. Detailed discussion and calculation of the average edit distance for each grammar is provided in~\cite{Comparison2018Wang}.

\begin{enumerate}[label=(\alph*), wide, labelwidth=!, labelindent=0pt]
  \item For grammar 1, 2 and 7, $D(G_{1,2,7}) = \infty$; 
  \item For grammar 3 and 4, $D(G_{3,4}) > 1$;
  \item For grammar 5 and 6, $D(G_{5,6}) = 1$.
\end{enumerate}

\section{Rule Extraction for Recurrent Networks}
\label{sec:extraction}
Rule extraction for recurrent networks essentially describes the process of developing or finding a rule that approximates the behaviors of a target RNN~\cite{jacobsson2005rule}. More formally, given a RNN denoted as a function $f: \mathcal{X} \rightarrow \mathcal{Y}$ where $\mathcal{X}$ is the data space, $\mathcal{Y}$ is the target space, and a data set $B = \{X, Y\}$ with $n$ samples $X \in \mathcal{X}^n$ and $Y \in \mathcal{Y}^{n}$. Let $r$ denote a rule which is also a function with its data and target space identical to that of $f$. The rule extraction problem is to find a function $\mathscr{L} : (\mathcal{X} \rightarrow \mathcal{Y}) \times (\mathcal{X}^n \times \mathcal{Y}^n) \rightarrow (\mathcal{X} \rightarrow \mathcal{Y}) $ such that $\mathscr{L}$ takes as input $f$ and $B$, then outputs a rule $r$. 

There are three key components in the above formulation - the extraction algorithm $\mathscr{L}$, a recurrent network $f$, and the underlying data sets $B$. In our previous study~\cite{Empirical2017Wang,Comparison2018Wang}, we investigated each component for their effect on the performance DFA extraction. More specifically, we empirically studied that when applying a quantization-based rule extraction algorithm to a second-order RNN~\cite{giles1990higher}, what conditions will affect DFA extraction and how sensitive DFA extraction is with respect to these conditions~\cite{Empirical2017Wang}. With respect to this question, we are interested in uncovering the relationship between different conditions. For instance, what is the influence of the initial condition of the RNN's hidden layer and the configuration of a particular quantization algorithm on DFA extraction. Specially, through our empirical study, we address the concerns of~\cite{kolen1994fool} by showing that DFA extraction is very insensitive to the initial conditions of the hidden layer. 

In addition, we also investigate how DFA extraction will be affected when we apply it to different recurrent networks trained on data sets with different levels of complexity. More specifically, when the underlying data sets are generated by Tomita grammars, we denote by $B_G$ a data set generated by a grammar $G$. Then in our evaluation framework, we fix the extraction method $\mathscr{L}$ as a quantization-based method and evaluate the performance obtained by $\mathscr{L}$ when its input, i.e. $B_G$ and $f$ trained on $B_{G}^{train}$~\footnote{Data set $B_G$ is split into a training set $B_{G}^{train}$ and a test set $B_{G}^{test}$ as typically done for supervised learning.}, vary across different grammars and different recurrent networks respectively. It is important to note that by comparing the extraction performance obtained by a given model across different grammars, we then examine for DFA extraction, how sensitive each model is with respect to the underlying data.

In the following, we introduce the rule extraction algorithm adopted in our previous work and the metrics proposed to evaluate the performance of DFA extraction.

\subsection{Quantization-Based DFA Extraction}
\label{sec:recipe}
Quantization-based DFA extraction methods have been the most frequently used in previous work~\cite{jacobsson2005rule,zeng1993learning,schellhammer1998knowledge,giles1992learning,omlin2000symbolic}. In these methods rules are constructed based on the hidden layers -- ensembles of hidden neurons -- of a RNN, and are also referred to as compositional approaches~\cite{jacobsson2005rule}. Also, it is commonly assumed that the vector space of a RNN's hidden layer can be approximated by a finite set of discrete states, where each rule refers to the transitions between states. As such, a generic compositional approach can be described by the following basic steps:
\begin{enumerate}[wide, labelwidth=!, labelindent=0pt]
\item Given a trained RNN, collect the values of a RNN's hidden layers when processing every sequence at every time step. Then quantize the collected hidden values into different states. This quantization is usually implemented with clustering methods. One such method that has been widely adopted is k-means clustering~\cite{zeng1993learning,frasconi1996representation,schellhammer1998knowledge,Empirical2017Wang}. In this study, we also use k-means due to its simplicity and computational efficiency.
\item Then use the quantized states and the alphabet-labeled arcs that connect these states to construct a transition diagram. Here we follow~\cite{schellhammer1998knowledge,Empirical2017Wang} and count the number of transitions observed between states. Then we only preserve the more frequently observed transitions.
\item Next, reduce the diagram to a minimal representation of state transitions with a standard and efficient DFA minimization algorithm~\cite{hopcroft2006automata} which has been broadly adopted in previous work for minimizing DFAs extracted from different recurrent networks and for other DFA minimization.
\end{enumerate}

There are other DFA extraction approaches, e.g., pedagogical approaches which construct rules by regarding the target RNN as a black box and build a DFA by only querying the outputs of this RNN for certain inputs. These approaches can be effectively applied to regular languages with small alphabet sizes. However, for RNNs which perform complicated analysis when processing sophisticated data, the extraction process becomes extremely slow~\cite{pmlr-v80-weiss18a}. This survey~\cite{jacobsson2005rule}  has a more detailed introduction of various rule extraction methods.

\subsection{Evaluation Metrics for DFA Extraction}
\label{sec:extract_metrics}
Here, we evaluate the performance of DFA extraction by measuring the quality of extracted DFAs. To be more specific, we introduce three metrics: (1)~the accuracy of an extracted DFA when it is tested on the test set for a particular grammar; (2)~the success rate from different random trials of extracting DFAs that are identical to the ground truth DFA associated with a particular grammar, which should then perform perfectly on the test set generated by that grammar; (3)~the fidelity of an extracted DFA from its source RNN when evaluated on the test set for a particular grammar. These metrics quantitatively measure the abilities of different recurrent networks for learning different grammars. In particular, the first metric reflects the abilities of different recurrent networks for learning ``good'' DFAs, and has been frequently adopted in much research~\cite{Ribeiro0G16,Hinton17Distilling,pmlr-v80-weiss18a}. The second metric, which is more rigorous, reflects the abilities of these models to learn correct DFAs. The third metric describes how similar an extracted DFA behaves with respect to the RNN from which the DFA is extracted. In the following, we formally introduce these metrics. It is important to note that our evaluation framework is agnostic to the underlying extraction method since we impose no constraints on $\mathscr{L}$. 

Given a model $m$ (which can either be a RNN $f$ or a DFA $r$) and a data set $B = \{X, Y\}$ consisting of samples $x$ and their corresponding labels $y_x$. Let $X_i$ denote the set of samples with the same label $i$, i.e., $X_i = \{ x \in X \mid y_x = i\}$. Then $X$ can be decomposed into disjoint subsets $X_i$. Similarly, let $X_i^m$ denote the set of samples classified by $m$ as having the label $i$, i.e., $X_i^m = \{ x \in X \mid m(x) = i\}$. Then we have the following metrics for evaluating the performance of DFA extraction.

\paragraph{Accuracy.} The accuracy $A(m, B)$ of model $m$ on data set $B$ is defined as :  
\begin{equation}
  \begin{aligned}
  \label{eq:acc}
  A(m, B) = \frac{1}{\left |  X\right |}\sum_{i=1}^{|\mathcal{Y}|} |X_i^m \cap X_i|,
  \end{aligned}
\end{equation}
where $|\cdot|$ denotes the cardinality of a set. To evaluate the accuracy of an extracted DFA $f$ on regular strings, we use $A(r, B) = \sum_{i=1}^{2} |X_i^r \cap X_i|/ |X|$.

\paragraph{Rate of Success.} The rate of success $S(r, B, T)$ of DFA extraction on data set $B$ over $T$ trails is defined as:
\begin{equation}
  \begin{aligned}
  \label{eq:success_rate}
  S(r, B, T) = \frac{1}{T}\sum_{t=1}^{T} \mathds{1}_{A(r_t, B) = 1},
  \end{aligned}
\end{equation}
where $r_t$ is the DFA extracted in the $t$-th trial. Correspondingly, the average accuracy of extracted DFAs is calculated by averaging $A(r_t, B)$ over $T$ trials.

\paragraph{Fidelity.} The fidelity $F(m_1, m_2, X)$ of two models $m_1$ and $m_2$ on a data set $X$ is defined as:
\begin{equation}
  \begin{aligned}
  \label{eq:fidelity}
  F(m_1, m_2, X) = \frac{1}{\left |  X\right |}\sum_{i=1}^{|\mathcal{Y}|} |X_i^{m_1} \cap X_i^{m_2}|.
  \end{aligned}
\end{equation}
Let $F(f, r, X)$ denote the fidelity of an extracted DFA $r$ regarding its source RNN $f$ on $X$, it is easy to derive that $F(f, r, X) = 1 - |X_1^{f} \bigtriangleup X_1^{r}| / |X|$. Here $X_1^f$ and $X_1^r$ denote the sets of strings classified as positive by $f$ and $r$, respectively, and $\triangle$ denotes the symmetric difference of two sets. 

In the following section, these three metrics are used to evaluate the DFA extraction performance for various recurrent networks.

\section{Experiments}
\label{sec:exp}
Here we present the experiment results of investigating the effect of various conditions (shown in Table~\ref{tab:factors}) on DFA extraction performance and verifying adversarial accuracy on different recurrent models. We first demonstrate that rule extraction performance is relatively stable for second-order RNN regardless of several varying conditions. Next, we show the evaluation results when we apply DFA extraction to different types of recurrent networks trained on data sets with different levels of complexity. Then, we present the results of verifying recurrent networks with DFAs.

\begin{table}[t]
\small
\begin{tabular}{
>{\columncolor[HTML]{EFEFEF}}c l}
\hline \hline
\multicolumn{1}{c|}{\cellcolor[HTML]{EFEFEF}Conditions}              & \multicolumn{1}{c}{Description}                                                           \\ \hline
\multicolumn{1}{c|}{\cellcolor[HTML]{EFEFEF}Data Complexity}         & Complexity of Tomita grammars                                                             \\ \hline
\multicolumn{1}{c|}{\cellcolor[HTML]{EFEFEF}}                        & \begin{tabular}[c]{@{}l@{}} $\circ$ Elman RNN, Second-order RNN, \\ \quad MI-RNN, GRU, LSTM\end{tabular} \\
\multicolumn{1}{c|}{\cellcolor[HTML]{EFEFEF}}                        & $\circ$ Randomly initialized hidden activation                                                    \\
\multicolumn{1}{c|}{\cellcolor[HTML]{EFEFEF}}                        & $\circ$ Size of the hidden layer                                                                  \\
\multicolumn{1}{c|}{\multirow{-4}{*}{\cellcolor[HTML]{EFEFEF}Model}} & $\circ$ Training epochs                                                                           \\ \hline
\multicolumn{1}{c|}{\cellcolor[HTML]{EFEFEF}Quantization}            & K for \emph{k}-means clustering                                                                  \\ \hline \hline
\end{tabular}
\caption{Conditions that affect DFA extraction.}
\label{tab:factors}
\end{table}

\subsection{Evaluation of DFA Extraction for Second-order RNN}
\label{sec:exp_2rnn}
Due to space constraints, we only present the extraction results of randomly initializing the hidden layer of second-order RNNs, and varying the pre-specified K for \emph{k}-means clustering. These two factors have been shown to be more influential than other conditions~\cite{Empirical2017Wang}, all shown in Table~\ref{tab:factors}. The extraction performance is evaluated by the average accuracy of extracted DFAs and the rate of success in DFA extraction. Discussion of the fidelity tests for second-order RNN and other recurrent networks is provided in Section~\ref{sec:exp_rnns}.

We followed~\cite{giles1992learning,Empirical2017Wang} and generated string sets by drawing strings from an oracle that generates random 0 and 1 strings for a grammar specified in Table~\ref{tab:tomita}. We verified each string from the random oracle and ensured they are not in the string set represented by that corresponding grammar before treating them as negative samples. It should be noticed that each grammar in our experiments represents one set of strings with unbounded size. As such we restricted the length of generated strings as previously specified~\cite{Empirical2017Wang}. We split the strings generated for each grammar to generate the training and test sets. Both data sets were used to train and test the RNNs accordingly, while only the test sets not used were used for evaluating extracted DFAs.
%%% see my changes above

We perform 130 trials of DFA extraction for each RNN on every grammar to comprehensively evaluate the performance of the DFA extraction. In particular, given a RNN and the data set generated by a grammar, we vary two factors -- the initial value of the hidden vector of this RNN (randomly initialized for 10 times.~\footnote{For each trial, we select a different seed for generating the initial hidden activations randomly.}) and the pre-specified value of $K$ for \emph{k}-means clustering in the range from 3 to 15. 

\begin{figure}[t]
\centering
\begin{subfigure}{.45\textwidth}
  \centering
  \includegraphics[width=\linewidth]{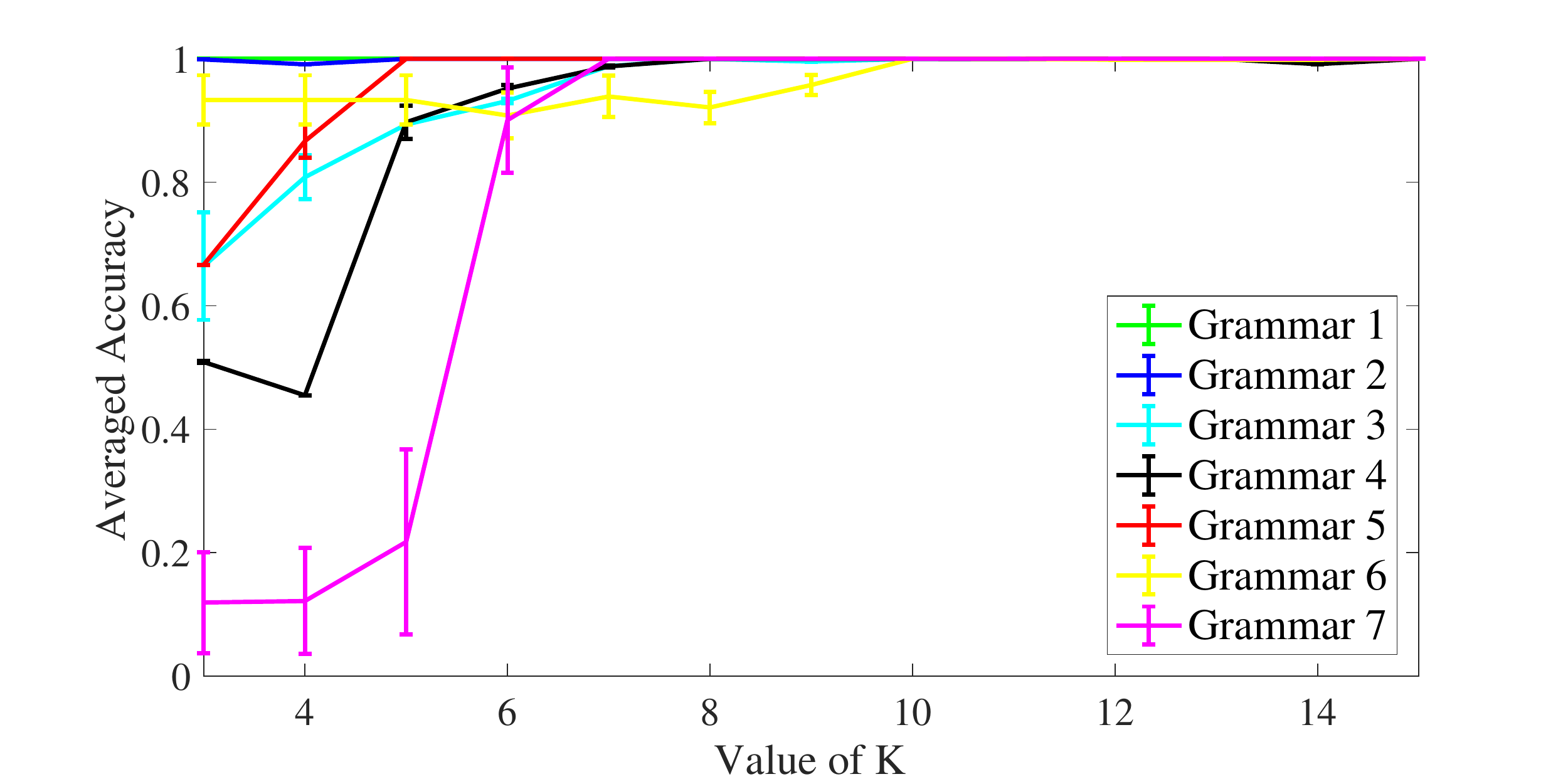}
  \centering
  %\vspace{-0.5em}
  \caption{Mean and variance of testing accuracy of extracted DFA with varying K for second-order RNN on the Tomita grammars~\cite{Empirical2017Wang}.}
  \label{fig:acc_2rnn}
\end{subfigure} \hfill
\begin{subfigure}{.45\textwidth}
  \centering
  \includegraphics[width=1.0\linewidth]{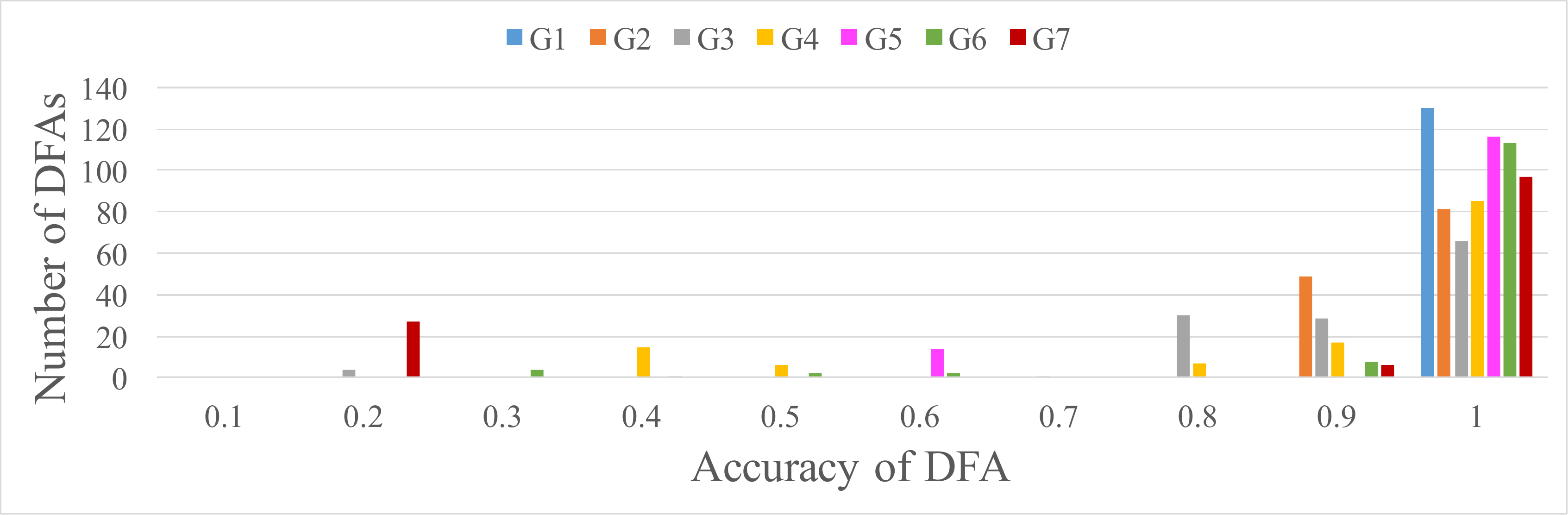}
  \centering
  \caption{Histograms of the classification accuracy of extracted DFAs for second-order RNN on the Tomita grammars~\cite{Empirical2017Wang}.}
  \label{fig:success_2rnn}
\end{subfigure} \hfill
\caption{Extraction performance for second-order RNN.}
\label{fig:exp_2rnn}
\end{figure}

\paragraph{Accuracy of Extracted DFA for Second-order RNN.}
As shown in Figure~\ref{fig:acc_2rnn}, for a sufficiently well trained (100.0\% accuracy on the test set) second-order RNN, the initial value of hidden layer has significant influence on the extraction performance when $k$ is set to small values. This impact can be gradually alleviated when $K$ increases. We observe that when $k$ is sufficiently large, the influence of randomly initializing the hidden layer is negligible.

\paragraph{Rate of Success for Second-order RNN.}
Besides showing the accuracy of the extracted DFAs, we further measure the success rate of extraction for second-order RNNs in Figure~\ref{fig:success_2rnn}. More specifically, the success rate of extraction is the percentage of DFAs with 100.0\% accuracy among all DFAs extracted for each grammar under different settings of $K$ and random initializations. From all 130 rounds of extraction for each grammar, we observe that the correct DFA successfully extracted with highest success rate of 100.0\% is on grammar 1, the lowest success rate of 50.0\% on grammar 3, and an averaged success rate of 75.0\% among all grammars. These results indicate that DFA extraction is relatively stable for a second-order RN on most grammars. 

\subsection{Evaluation of DFA Extraction for Different RNNs}
\label{sec:exp_rnns}

\begin{figure}[t]
\centering
\begin{subfigure}{.45\textwidth}
  \centering
  \includegraphics[width=\linewidth]{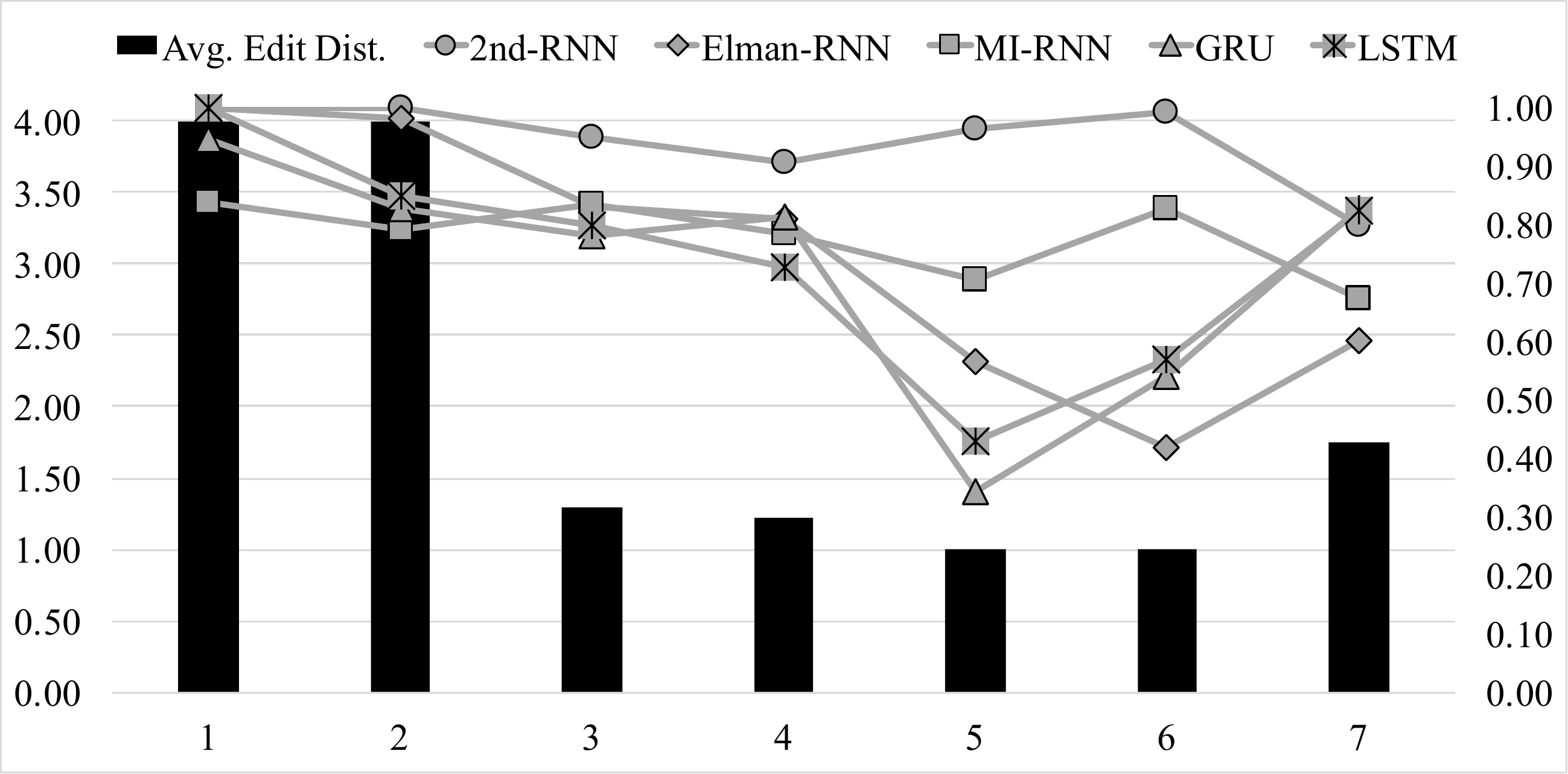}
  \centering
  %\vspace{-0.5em}
  \caption{Average accuracy of DFAs extracted from recurrent networks on the Tomita grammars. Left vertical axis: average edit distance. Right vertical axis: average accuracy of extracted DFAs~\cite{Comparison2018Wang}.}
  \label{fig:acc_dist}
\end{subfigure} \hfill
\begin{subfigure}{.45\textwidth}
  \centering
  \includegraphics[width=1.0\linewidth]{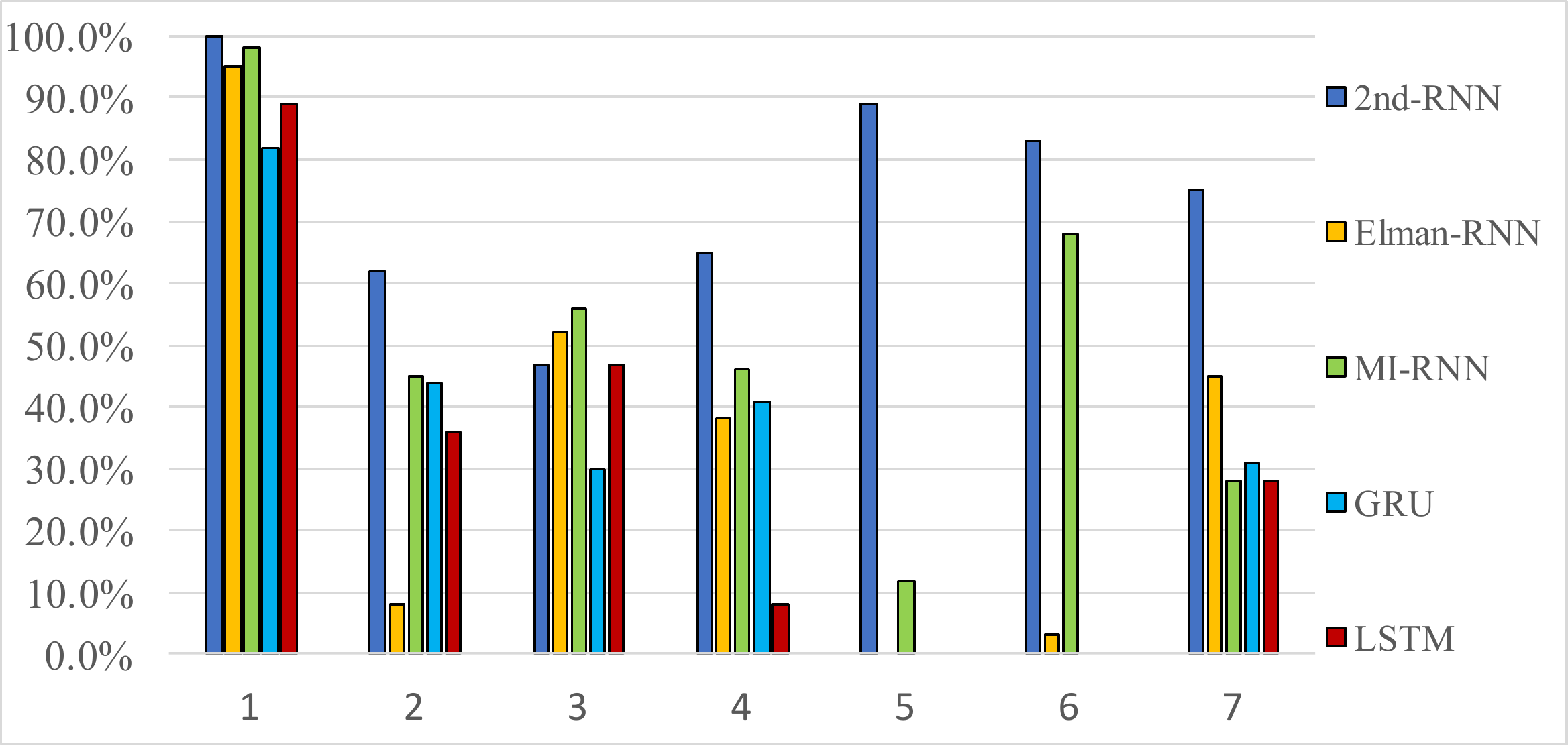}
  \centering
  \caption{Histograms of the classification accuracy of extracted DFAs for different RNNs on all grammars~\cite{Comparison2018Wang}.}
  \label{fig:success_all}
\end{subfigure} \hfill
\caption{Extraction performance for different RNNs.}
\label{fig:exp_all_rnn}
\end{figure}

Though we empirically investigated DFA extraction for second-order RNNs, it is not clear if DFA extraction can be effectively applied to different recurrent networks, and what is the cause for the inconsistent extraction performance observed across different grammars. To address these questions, we empirically evaluate DFA extraction performance for other recurrent networks 
%besides second-order RNN~\cite{giles1992learning} 
for Elman RNN~\cite{elman1990finding}, multiplicative integration recurrent neuron networks (MI-RNN)~\cite{wu2016multiplicative}, long-short-term-memory networks (LSTM)~\cite{hochreiter1997long} and gated-recurrent-unit networks (GRU)~\cite{cho2014properties}. We also show in Figure~\ref{fig:acc_dist} that the complexity of different Tomita grammars is the underlying reason for the inconsistent extraction performance.

Our experiment setup is the same as that for second-order RNNs. In particular, for every pair of a recurrent network and a grammar, we conducted 10 trials with random initialization of the hidden layer of that RNN, and apply DFA extraction for this RNN multiple times by ranging $K$ from 3 to 15. We tested and recorded the accuracy of each extracted DFA using the same test set constructed for evaluating all corresponding recurrent networks. The extraction performance is then evaluated based on results obtained from these trials. This we believe alleviates the impact of different recurrent networks being sensitive to certain initial state settings and clustering configurations. Also, we used recurrent networks with approximately the same number of weight and bias parameters regardless of their different architectures.
%% see my edits
\paragraph{Accuracy of Extracted DFA for Different RNNs.}
In Figure~\ref{fig:acc_dist}, we plot the average accuracy of 130 DFAs extracted from each model trained on each grammar, and the average edit distance of each grammar calculated by setting $N = 20$. As shown in Figure~\ref{fig:acc_dist}, except for second-order RNN and MI-RNN, the average accuracy obtained by DFAs extracted from each model decreases as the average edit distance of grammars decreases. This indicates that it is generally more difficult for recurrent networks to learn a grammar with a higher level of complexity. 

\paragraph{Rate of Success for Different RNNs.}
The results for evaluating and comparing different RNN models on their rate of success in extracting the correct DFAs associated with the Tomita grammars are shown in Figure~\ref{fig:success_all}. We find that on grammars with lower complexity, all models are capable of producing the correct DFAs. In particular, all models achieve much higher success rates on grammar 1. This may due to the fact that the DFA associated with grammar 1 has the fewest number of states (two states) and simplest state transitions among all other DFAs. Thus, the hidden vector space for all RNN models is much easier to separate during training and identify during extraction. As for other grammars with lower complexity, their associated DFAs have both a larger number of states and more complicated state transitions. For grammars with higher levels of complexity, the second-order RNN enables a much more accurate and stable DFA extraction. Also, for the most part, the MI-RNN provides the second best extraction performance, especially, for grammars 5 and 6, which have the highest complexity. In this case only the second-order RNN and MI-RNN are able to extract correct DFAs, while all other models fail.

\paragraph{Fidelity of Extracted DFAs for Different RNNs.}

\begin{table}[t]
\small
\begin{tabular}{c|l|cc|c}
\hline \hline
\rowcolor[HTML]{C0C0C0} 
\cellcolor[HTML]{C0C0C0}                          & \multicolumn{1}{c|}{\cellcolor[HTML]{C0C0C0}}                        & \multicolumn{2}{c|}{\cellcolor[HTML]{C0C0C0}RNN Evaluation} & \cellcolor[HTML]{C0C0C0}                           \\
\rowcolor[HTML]{C0C0C0} 
\multirow{-2}{*}{\cellcolor[HTML]{C0C0C0}Grammar} & \multicolumn{1}{c|}{\multirow{-2}{*}{\cellcolor[HTML]{C0C0C0}Model}} & Clean                        & Noisy                        & \multirow{-2}{*}{\cellcolor[HTML]{C0C0C0}Fidelity} \\ \hline \hline
                                                  & 2nd-RNN                                                              & 1.00                         & 0.99                         & 1.00                                               \\
                                                  & Elman-RNN                                                            & 1.00                         & 0.99                         & 1.00                                               \\
                                                  & \textbf{MI-RNN}                                                               & \textbf{1.00}                         & \textbf{0.99}                         & \textbf{0.98}                                               \\
                                                  & GRU                                                                  & 1.00                         & 0.99                         & 1.00                                               \\
\multirow{-5}{*}{3}                               & LSTM                                                                 & 1.00                         & 0.99                         & 1.00                                               \\ \hline \hline
                                                  & 2nd-RNN                                                              & 1.00                         & 0.99                         & 0.99                                               \\
                                                  & Elman-RNN                                                            & 1.00                         & 0.99                         & 0.93                                               \\
                                                  & \textbf{MI-RNN}                                                               & \textbf{0.99}                         & \textbf{0.99}                         & \textbf{0.69}                                               \\
                                                  & GRU                                                                  & 0.99                         & 0.99                         & 0.99                                               \\
\multirow{-5}{*}{4}                               & LSTM                                                                 & 1.00                         & 0.99                         & 0.89                                               \\ \hline \hline
                                                  & 2nd-RNN                                                              & 1.00                         & 0.99                         & 1.00                                               \\
                                                  & Elman-RNN                                                            & 1.00                         & 0.99                         & 1.00                                               \\
                                                  & \textbf{MI-RNN}                                                               & \textbf{0.99}                         & \textbf{0.99}                         & \textbf{0.40}                                               \\
                                                  & GRU                                                                  & 1.00                         & 0.99                         & 0.99                                               \\
\multirow{-5}{*}{7}                               & LSTM                                                                 & 1.00                         & 0.99                         & 1.00                                               \\ \hline \hline
\end{tabular}
\caption{Fidelity test results for different recurrent networks on grammar 3, 4 and 7. Columns ``Noisy'' and ``Clean'' present the results for evaluating RNNs on the data sets with and without noise, respectively.}
\label{tab:fidelity}
\end{table}

\begin{figure}[t]
\hfill
\begin{center}
\includegraphics[width=0.95\linewidth]{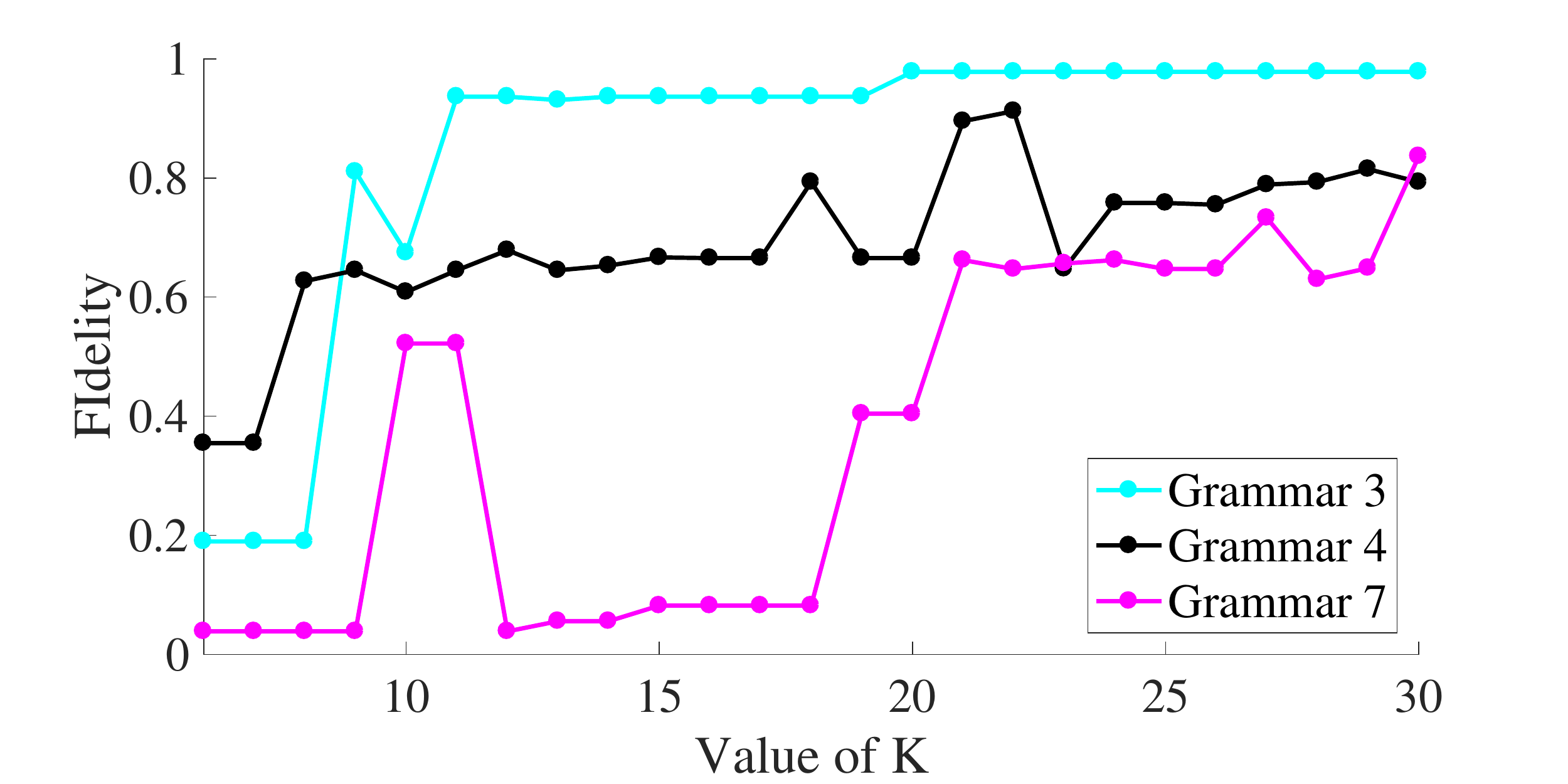}
\end{center}
\caption{Fidelity test of MI-RNN regarding K (for \emph{k}-means) ranging from 6 to 30 on grammar 3, 4 and 7.}
\label{fig:fidelity_mirnn}
\end{figure}

When both a RNN and the DFA extracted from that particular RNN obtain 100\% accuracy on the test set, it is trivial to check that the fidelity has a value of 1. In order to evaluate the fidelity in a more realistic scenario, we inject noise into the training sets by randomly selecting several training samples and flipping their labels. To avoid causing a RNN to be severely biased by this noise, we limit the number of samples selected to 4, i.e., 2 for positive strings and 2 for negative strings. We then train a RNN on the noisy training set and extract DFAs accordingly. The fidelity is calculated as discussed in Section~\ref{sec:extract_metrics}. Table~\ref{tab:fidelity} shows the results for fidelity tests for all recurrent networks on grammars 3, 4 and 7 from a single trial.~\footnote{Other grammars are not shown due to space constraints.} Specifically, during the extraction, we set the value of $K$ to 20 since as shown in Figure~\ref{fig:acc_2rnn} a larger $K$ is more likely to provide an accurate DFA. As shown in Table~\ref{tab:fidelity}, for most recurrent networks the high fidelity values obtained by the extracted DFAs from these models across three grammars indicate that these networks can effectively tolerate training set noise. An exception is MI-RNN, from which the extracted DFAs have consistently the lowest fidelity values across three grammars. This shows that MI-RNN is more sensitive to the training set noise. This results in reducing its overall classification performance and causing worse extraction performance. To better illustrate this effect, we show in Figure~\ref{fig:fidelity_mirnn} how the fidelity varies as we increase the value of $K$ from 6 to 30. As $K$ increases, the DFAs extracted from MI-RNN on these grammars have better accuracy on the clean data sets. As such, the similarity between extracted DFAs and their source RNNs increases. This result indicates that a proper setting of $K$ is important for DFA extraction that is both accurate and faithful.

In general, Elman-RNN obtains the worst extraction performance on most grammars, while DFAs extracted from second-order RNN and MI-RNN have consistently higher accuracy and rate of success across all grammars. The former may due to the simple recurrent architecture of Elman-RNN, which possibly limits its ability to fully capture complicated symbolic knowledge. The better extraction performance of second-order RNN and MI-RNN raises questions regarding the quadratic interaction between input and hidden layers used by these models and whether such an interaction could improve other models' DFA extraction, an interesting question for future work.

\subsection{Verification of Recurrent Networks with DFAs}
\label{sec:exp_verify}
Following the verification framework described in Section~\ref{sec:overview}, we present in the following experiment the results of verifying recurrent networks with DFAs. It is important to note that when selecting a ground truth DFA as the oracle, we can comprehensively examine the robustness or adversarial accuracy~\cite{tjeng2017evaluating} of a certain RNN with respect to small-scale perturbations. If we select an extracted DFA as the oracle, then our verification framework can be adopted for examining the fidelity of the extracted DFA. The latter case has been demonstrated with a simplified case study shown in the previous section, here we focus on the former case.

\begin{algorithm}[t]
\caption{Adversarial Accuracy Verification}\label{algm: brute_force}
\begin{algorithmic}[1]

\REQUIRE RNN $f$; Extracted DFA $r$; String length $N$;\\ Number of samples $T$; Allowed perturbed distance $d$;
\ENSURE Adversarial accuracy $\gamma$;\\

\STATE Randomly generate $T$ samples $X$ with length $N$, and $X=\{x_i \mid r(x_i)=f(x_i)=p\}$, where $p$ denotes the positive label;
\STATE $count \gets 0$;

\FOR{$i=1$ to $T$} 
\STATE Generate samples $O_{x_i}$ from $x_i$ satisfying $O_{x_i}=\{ x_{j} \mid d_{e}(x_i,x_{j})\leq d)\}$;
\FOR{$j=1$ to $\left | O_{x_i}\right |$} 
\IF{$r(x_{j})=n$}
\STATE \textbf{Continue};
\ELSIF {$f(x_{ij})=n$ (where $n$ denotes the negative label)}
\STATE $count \gets count+1$;
\STATE \textbf{Break};
\ENDIF
    
\ENDFOR
\ENDFOR

\STATE $\gamma \gets 1 - count/N$;
\RETURN $\gamma$;
\end{algorithmic}
\end{algorithm}

Given a well trained RNN and a ground truth DFA associated with the grammar used for training this RNN, our verification task mainly focuses on the local invariance property~\cite{leofante2018automated} of the RNN. More specifically, we verify the case when a small-scale perturbation is applied to a positive string $x$, whether a RNN will produce a negative label while the DFA still classifies $x$ as positive~\footnote{We also report the results for verifying the local invariance property of a RNN with negative strings.}. Here we only use grammar 3, 4 and 7 for the verification task. This is because for other grammars, it is easy to check that given a positive string $x$, almost all strings with the edit distance to $x$ equals $1$ belong to the negative class. This means that for grammar 1, 2, 5 and 6, all their positive samples lie on the decision boundary hence the perturbation space is rather limited. While for grammar 3, 4 and 7, it is easier to find adversarial samples with small perturbed edit distance. As such, in the following experiments, we set the maximal allowed perturbed edit distance as 1 to satisfy the constraint mentioned in Section~\ref{sec:overview}.

Since all recurrent networks have been sufficiently well trained on short strings that make up the training and test sets, we verified these models' adversarial accuracy on long strings. It is well known that recurrent networks have difficulty capturing long-term dependencies. As such, we randomly sampled strings with length 200 to construct the verification data sets. All sampled strings were examined to be correctly classified by both a target recurrent network and the ground truth DFA for grammar 3, 4 and 7. Since the number of strings increases exponentially as their length increases, we randomly sampled 100 positive and 100 negative strings for 30 trials for verification. This allows us to better approximate the ideal results by exploring the entire data space. Based on the verification framework introduced in Section~\ref{sec:overview}, we design the verification algorithm for a single trial as shown in Algorithm~\ref{algm: brute_force}. 

\begin{figure}[t]
\hfill
\begin{center}
\includegraphics[width=0.95\linewidth]{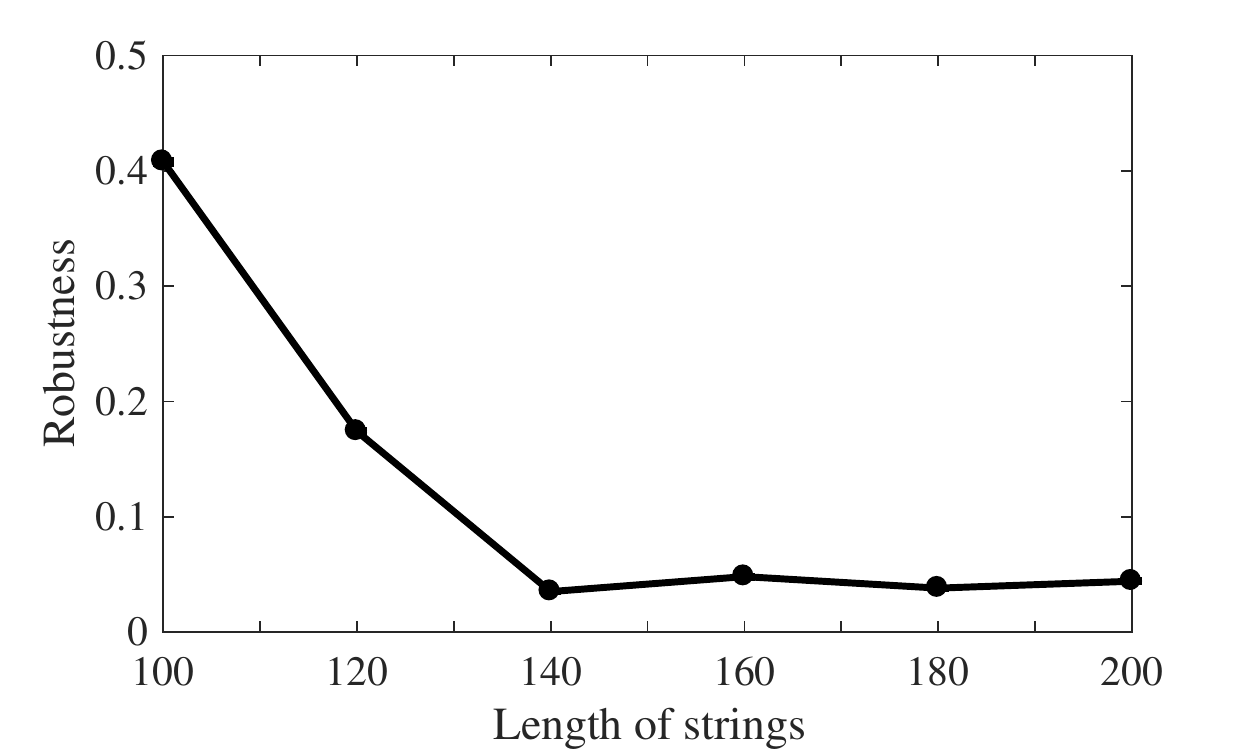}
\end{center}
\caption{Adversarial accuracy of an Elman-RNN for strings with varying length (from 100 to 200) on grammar 3.}
\label{fig:robust_vary_len}
\end{figure}

The $\gamma$ obtained from 30 trials of verifying positive (negative) strings are averaged and denoted as $\bar{\gamma}_{+}$ ($\bar{\gamma}_{-}$). The results presented in Table~\ref{tab:verify_results} indicate the different levels of robustness obtained by different recurrent networks. In particular, second-order RNN and MI-RNN are most robust with no adversarial samples identified, while other recurrent networks suffer from adversarial samples to a different extent. These results are consistent with the results reported previously in Section~\ref{sec:exp_rnns} and~\cite{Comparison2018Wang}. Of note, the Elman-RNN obtains the lowest adversarial accuracy when verified for positive strings from grammar 3. To understand the reason for this worst result, we show in Figure~\ref{fig:robust_vary_len} how the adversarial accuracy of an Elman-RNN changes when for verification the length of strings sampled changes. This indicates that an Elman-RNN cannot generalize to long strings and may cause it to more likely suffer from adversarial attacks.

\begin{table}[t]
\small
\begin{tabular}{ccc|ccccc|}
\cline{4-8}
\multicolumn{3}{l}{}                                                                                 & \multicolumn{5}{c|}{\cellcolor[HTML]{C0C0C0}RNN} \\ \hline
\rowcolor[HTML]{C0C0C0} 
\multicolumn{1}{|c|}{\cellcolor[HTML]{C0C0C0}G} & \multicolumn{1}{c|}{\cellcolor[HTML]{C0C0C0}$y$} & $\gamma$ & 2nd    & Elman      & MI-RNN   & GRU    & LSTM \\ \hline \hline
\multicolumn{1}{|c|}{}                          & \multicolumn{1}{c|}{1}                         & $\bar{\gamma}_{+}$ & 1.00 & \textbf{3.96e-2}  & 1.00 & 1.00 & 1.00     \\
\multicolumn{1}{|c|}{\multirow{-2}{*}{3}}       & \multicolumn{1}{c|}{0}                         & $\bar{\gamma}_{-}$ & 1.00 & 1.00 & 1.00       & 1.00 & \textbf{0.96}   \\ \hline \hline
\multicolumn{1}{|c|}{}                          & \multicolumn{1}{c|}{1}                         & $\bar{\gamma}_{+}$ & 1.00 & 1.00 & 1.00       & 1.00 & 1.00            \\
\multicolumn{1}{|c|}{\multirow{-2}{*}{4}}       & \multicolumn{1}{c|}{0}                         & $\bar{\gamma}_{-}$ & 1.00 & 1.00 & 1.00       & 1.00 & 1.00            \\ \hline \hline
\multicolumn{1}{|c|}{}                          & \multicolumn{1}{c|}{1}                         & $\bar{\gamma}_{+}$ & 1.00 & \textbf{0.99} & 1.00 & \textbf{0.99} & \textbf{0.98} \\
\multicolumn{1}{|c|}{\multirow{-2}{*}{7}}       & \multicolumn{1}{c|}{0}                         & $\bar{\gamma}_{-}$ & 1.00 & 1.00 & 1.00       & 1.00 & 1.00            \\ \hline
\end{tabular}
\caption{Verification results for positive and negative strings with the length of 200.}
\label{tab:verify_results}
\end{table}

It can also be observed from Table~\ref{tab:verify_results} that the difference between recurrent networks' robustness against adversarial samples may result from a difference between the underlying grammars. Specifically, grammar 4 enables better robust learning than grammar 3, even though these two grammars have similar levels of complexity. As for grammar 7, although it has the lowest complexity in comparison with grammar 3 and 4, there are effective adversarial samples identified for most recurrent networks. This indicates that this grammar is prone to overfitting the recurrent networks since the data sets for this grammar is very imbalanced for positive and negative samples.

\section{Conclusions and Discussion}
Here we propose to verify recurrent networks with DFA extraction. We extend the verification framework proposed in prior work for feed-forward neural networks to accommodate what the rigorous requirements for verification of recurrent networks. In particular, we empirically study DFA extraction on various recurrent networks. We show that for certain recurrent networks, their extracted DFAs have such an accuracy that they can be regarded as surrogates for their ground truth counterparts to be used in the verification task. We also show through a case study that our verification framework can also be adopted for examining the equivalence between an extracted DFA and its source RNN using a fidelity metric. In addition, we define an average edit distance metric that is suitable for measuring the adversarial perturbation applied to strings generated by regular grammars. These results are then used in an experimental study of verification for several different recurrent networks. The results demonstrate that while all recurrent networks can sufficiently learn short strings generated by the different Tomita grammars, only certain RNN models can generalize to long strings without suffering from adversarial samples.

Future work would include employing a DFA-based verification for model refinement and conducting more efficient fidelity tests between an extracted DFA and the source recurrent network. Specifically, since a DFA is usually much easier for formal analysis, we could efficiently identify certain implicit weaknesses of a RNN by using a DFA extracted from that RNN to generate specific adversarial samples. The generated adversarial samples could then be used for refining the source RNN. In addition, as discussed in Section~\ref{sec:overview}, for DFA-based verification for a RNN, it is crucial to extract a DFA that is faithful to the source RNN. Indeed, this fidelity requirement is critical for not only verification but also for explanability. A comprehensive fidelity test can be very challenging for recurrent networks since the dimension of sequential data expands exponentially. This also raises a limitation on the computational efficiency of conducting verification for this work. This is largely due to the difficulty of computing edit distance, which results in solving an old NP-hard problem~\cite{backurs2015edit,de2010grammatical}. As such, future work could be in exploring more efficient approximation algorithms.
%%% need citation to NP-hard problem

\bibliography{ref}
\bibliographystyle{aaai}
\end{document}